\documentclass[runningheads]{llncs}
\usepackage[T1]{fontenc}
\usepackage{graphicx}
\usepackage{amsmath}
\usepackage{multirow}
\usepackage{hyperref}

\begin{document}
\title{Multimodal LLMs Struggle with Basic Visual Network Analysis: a VNA Benchmark}
%
%
\author{Evan M. Williams\inst{1}\orcidID{0000-0002-0534-9450} \and
Kathleen M. Carley\inst{1}}
\authorrunning{Williams and Carley}
%
\institute{Carnegie Mellon University, Pittsburgh, PA 15213, USA \\
\email{\{emwillia,carley\}@andrew.cmu.edu}}
\maketitle              
\begin{abstract}
We evaluate the zero-shot ability of GPT-4 and LLaVa to perform simple Visual Network Analysis (VNA) tasks on small-scale graphs. We evaluate the Vision Language Models (VLMs) on 5 tasks related to three foundational network science concepts: identifying nodes of maximal degree on a rendered graph, identifying whether signed triads are balanced or unbalanced, and counting components. The tasks are structured to be easy for a human who understands the underlying graph theoretic concepts, and can all be solved by counting the appropriate elements in graphs. We find that while GPT-4 consistently outperforms LLaVa, both models struggle with every visual network analysis task we propose. We publicly release the first benchmark for the evaluation of VLMs on foundational VNA tasks.  

\end{abstract}
\section{Introduction}

Large Language Models (LLMs) and Large Vision Language Models (VLMs) are transforming the ways people work and research. There has recently been a surge of interest in understanding how LLMs can be used in network analytic workflows and evaluating their performance on network-level tasks \cite{fatemi2023talk}. At the time this was first submitted to ArXiv, there were not yet any public visual network analysis tasks. However, two datasets have since been released; we note that none of the tasks introduced in this work were considered in the concurrent works or benchmarks. We introduce the broad task of zero-shot Visual Network Analysis (VNA), which we define as deriving graph theoretic concepts from network visualizations without relying on human-annotated examples. We introduce a new VNA benchmark and evaluate the performance of GPT-4 \cite{achiam2023gpt} and LLaVa1.5-3b \cite{liu2024llavanext} on 5 tasks based on 3 foundational network science concepts.

We introduce 5 VNA tasks based on three simple and important graph theory concepts: 1) degree, 2) structural balance, and 3) components. We create two tasks related to degree. Given a visualized graph, we ask the multimodal LLM to 1a) identify the maximum degree of the graph, i.e., the largest degree of any node, and 1b) to return the node IDs of all nodes with the maximum degree. We create one task based on structural balance: 2a) given an image of a triad with edges colored to denote a relation type, we ask the multimodal LLM to assess whether the triad is balanced or imbalanced. Finally, given an image of graphs with multiple components, 3a) we ask the LLM to count the number of components in the graph, and 3b) we ask the LLM to count the number of isolates.

Each of these tasks is related to an important graph theoretic concept, but each task is also connected in how it can be solved. Every task we create can be solved by counting the correct element within each graph. For the degree-level tasks, the multimodal LLM needs to count edges incident to the nodes that appear to have the largest number of edges. In the structural balance tasks, the answer can be deduced by either counting the number of positive or negative edges in the triad. The component tasks are very explicitly asking the multimodal LLM to count the number of components and the number of isolates. Consequently, these tasks, as we've constructed them, are highly related to zero-shot object counting \cite{xu2023zero}. For each of these tasks, we generate synthetic, high-resolution, graph visualizations while prioritizing human readability. We publish the all of the data generated for this project\footnote{\url{https://figshare.com/articles/dataset/Multimodal_LLMs_Struggle_with_Basic_Visual_Network_Analysis_a_Visual_Network_Analysis_Benchmark/25938448}} 

We find that GPT-4 and LLaVa struggle on all 5 tasks we propose. Across all experiments, the highest accuracy was achieved by gpt-4 on the isolate counting task, where it correctly identified the number of isolates in 67 of 100 graph visualizations. Predicting whether or not triads were structurally balanced was surprisingly one of the most challenging tasks for both LVMs, despite the simplicity of the task. The best performing model---gpt-4---achieved an overall accuracy of 0.51 on the task, on par with random guessing. More work is needed to understand and improve zero-shot LVM performance on visual graph analysis tasks.

\section{Related Works}

Recent work has explored potential Large Language Model (LLM) applications within social networks \cite{zeng2024large}, and in generating features and predictions for machine learning applications on graphs \cite{chen2024exploring}. Recent work has also explored how best to encode graphs as text for LLM analysis \cite{fatemi2023talk}. However, each of these works look solely at LLMs and do not consider VLM usage. Two days prior to the release of this publication on ArXiv, the VisionGraph Benchmark was released \cite{li2024visiongraph}. The visiongraph benchmark contains complementary tasks and evaluates VLM performance on many relatively-complex graph tasks, including cycle identification, identifying shortest paths, and identifying maximum flow \cite{li2024visiongraph}. Wei et al. concurrently introduced a GITQA (Graph-Image-Text Question Answering) Dataset and an end-to-end framework for general graph reasoning \cite{wei2024rendering}.

There has also been a recent surge in interest in zero-shot evaluation capabilities Large Vision Language Model (VLM) evaluations. These evaluations have been applied to a wide range of computer vision tasks \cite{zhang2024vision}. \cite{xu2023zero} propose the task of zero-shot object counting, which they define as counting the number of instances of a specific class in the input image given only the class name. Llava has been evaluated on various zero-shot tasks, and has performed well on differentiating animals, counting animals, identifying written digits, and identifying pox \cite{islam2023pushing}. Previous work has found that VLMs perform worse than specialized models at object counting, and seemed to perform better at counting cars than counting trees or animals \cite{zhang2024good}. 

\section{Methods}

In all tasks, each graph is independently sent to gpt-4 along with its accompanying text prompt using OpenAI's API. We also feed each graph, independently, to LLaVa, which we run on a local cluster. All LLaVa prompts were modified to include the suggested prompt format on which LLaVa was trained: "USER: <image>$\textbackslash$n< prompt >$\textbackslash$nASSISTANT:".

\subsection{Maximum Degree Tasks}

Degree centrality is likely the most widely-used graph centrality statistic. For an undirected graph $G$, the degree of node $i$ is simply the number of edges incident to node $i$. For a binary, symmetric, and undirected adjacency matrix $A$, degree is simply defined as $\sum_i^N x_i$. In social networks, degree centrality often corresponds to popularity or engagement, e.g., in a friendship network, a node with a high degree has many friends.

We consider two different prompts for each graph. In the first prompt, we ask the question using Graph Theory terminology---we call this our 'formal prompt'. In the second question, we attempt to ask the question in a more human way. We state that the graph is a first-grade friendship network and ask the same questions using the terminology like popular students and total number of friends---we call this the 'human prompt'. We consider two prompts for each graph. The first asks for the largest degree centrality in the graph and the list of all nodes with that centrality. The second prompt states that we are observing a first-grade friendship network and asks who has the most friends. We exclude exact prompts from this paper due to space constraints, but we make all of prompts publicly available online\footnote{\url{https://github.com/EvanUp/VNA_Benchmark/blob/main/prompts/prompts.csv}}. LLaVa would return blank fields if given the same formatting stipulations as GPT-4, so those were dropped in the LLaVa prompts. In the cases where LLaVa made contradictory statements or inaccurate statements, e.g., ``A has a degree centrality of 10, while B has a degree centrality of 9. The largest degree centrality is B, which is 9'', we selected the most logically coherent option---in this case, (A, 10). In cases where LLaVa or GPT-4 did not return a numeric maximum degree or node IDs, we impute a maximum degree of 0 and assign the response an empty set of node IDs. 



\subsection{Structural Balance Task}

Balance Theory, which dates back to Fritz Heider's 1946 studies of individual cognition and perception of social situations, is a core concept in social network analysis \cite{heider1946attitudes}. Heider's original formalization of \textit{cognitive balance} was generalized in the 1950s to \textit{structural balance}, which focuses on a group of people rather than an individual \cite{cartwright1956structural,harary1953notion}. Given a undirected signed graph $\mathcal{G}^{+}_{-}$, where each edge can assume a value of ``like'' ($+$) or ``dislike'' ($-$), structural balance occurs when $i$ likes $j$ ($i \stackrel{+}\leftrightarrow j$) and $i$ and $j$ agree in their evaluation of $k$, i.e., $(i \stackrel{+}\leftrightarrow k \land j \stackrel{+}\leftrightarrow k) 	\lor (i \stackrel{-}\leftrightarrow k \land j \stackrel{-}\leftrightarrow k) $ . Concretely, for signed, undirected, triads, this means that any triad with a total of 1 or 3 ($+$) edges is balanced. Conversely, triads with 0 or 2 ($+$) edges are considered unbalanced. Consequently, of the 8 possible signed triads, 4 are structurally balanced and 4 are unbalanced. For more information on structural balance we refer the reader to chapter 6 of \cite{wasserman1994social}. We considered two different prompts, one without a definition of structural balance and one containing a simple definition of structural balance. 

\subsection{Component Tasks}

A component is a connected graph or subgraph that is not a part of a larger connected subgraph. In undirected graphs, the number of components can be defined as the minimum number of random walkers that it would take to reach every node on a graph. Nodes of degree zero are a special type of component commonly referred to as isolates. We considered two prompts for this task---one without definitions and one with definitions. Again, LLaVa was not generating responses with the formatting requirements included in the prompt, so its prompt was truncated. LLaVa's return format with this prompt was largely consistent and answers could be reliably extracted with a regex. Of the 400 LVM calls for this task, there were 4 instances, all from GPT-4, where no numbers were returned. These instances were imputed with zeros. 

\section{Data}

For each experiment, we programatically generate graphs using the python libraries NetworkX and netgraph. For the degree tasks, graphs were generated with Kamada Kawai layouts, as we qualitatively found these to be the most human-readable for visual node degree tasks. We bolded font weights in the degree and structural balance tasks for the same reason. All graphs are exported as high-quality 300 DPI .png files. 

\subsection{Maximum Degree Graph Generation}

\begin{figure}[!h]
    \centering
    \includegraphics[scale=0.5]{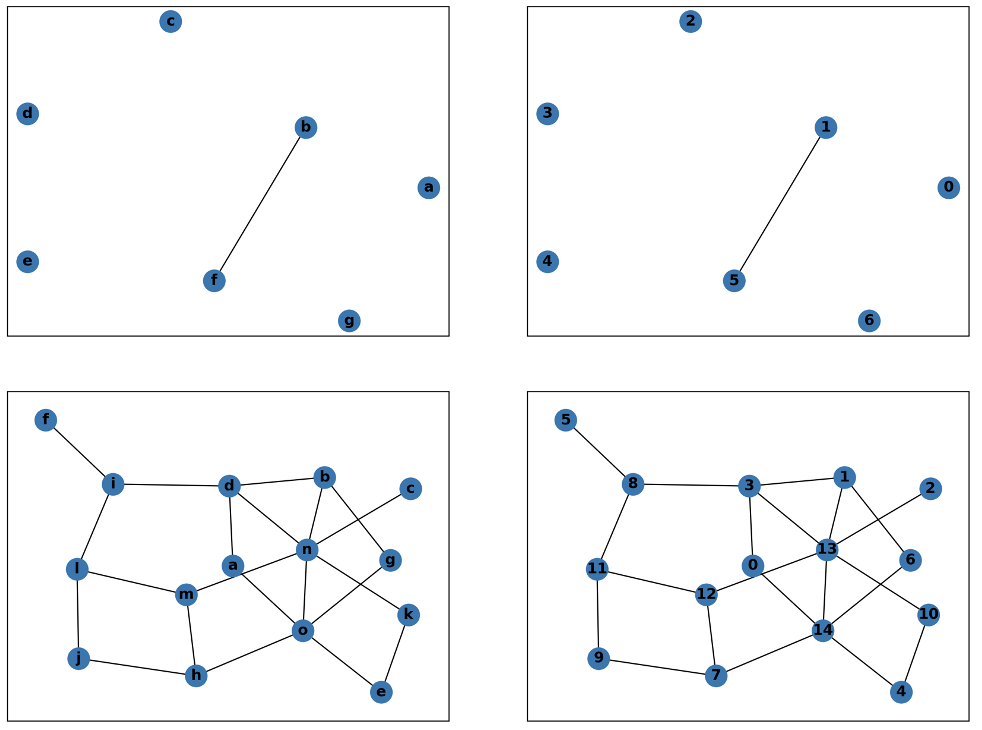}
    \caption{Degree Task Graph Examples with letter (left) and numeric (right) node IDs.}
    \label{fig:degree}
\end{figure}

We provide the LLMs with two degree-related tasks. First, we ask the LLM to find the maximum degree of the graph. Second, we ask the LLM to return the node IDs of all nodes with the maximum degree. In large graphs, this can be a challenging, or even infeasible, task for humans. Consequently, we consider only relatively-sparse graphs ranging from 1 to 20 nodes. We generate 20 Erdos-Renyi graphs, each with the parameter $p$ set to $0.2$. As LVMs have previously been found to be prone to typographic attacks \cite{goh2021multimodal}, we considered that numeric node IDs could impact the ability of LLMs to return numbers. Consequently, we generate two identical versions of each graph: one with numeric node IDs and one with alphabetical node IDs (see Figure \ref{fig:degree}). 

\subsection{Structural Balance Graph Generation}

\begin{figure}[!h]
    \centering
    \includegraphics[scale=0.25]{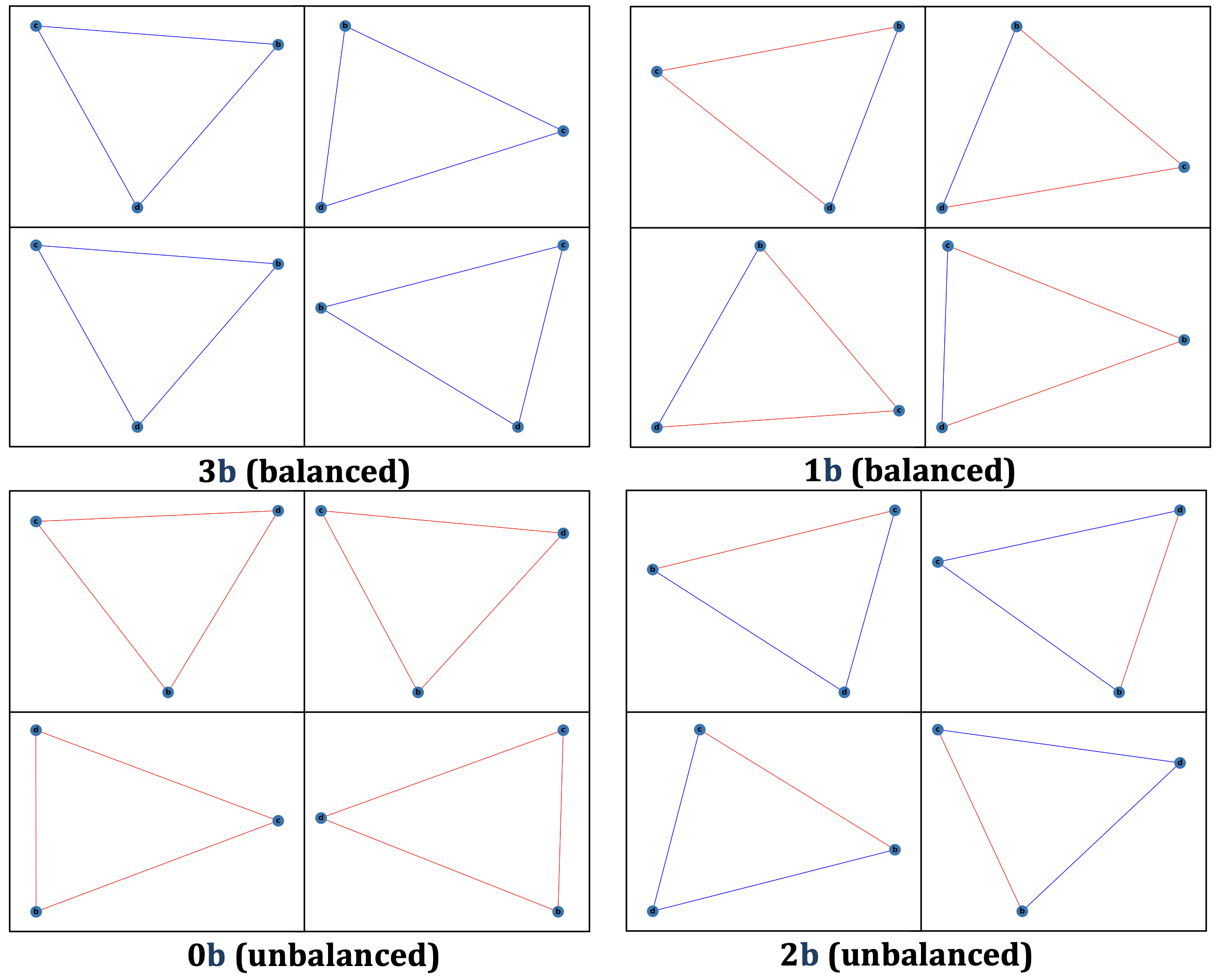}
    \caption{Triadic Balance Examples. Top row contains  a sample of balanced triads, bottom row contains a sample of unbalanced triads. 'b' denotes the number of like (blue) relationships in each group.}
    \label{fig:triads}
\end{figure}

For each of the 8 possible signed triads, we generate 10 graphs, each with random layouts for a total of 80 triad graphs. In each graph, "like" edge relations are colored blue and "dislike" edge relations are colored red. As node IDs are unimportant for this task, we arbitrarily chose to use letter node IDs. We further group types of triads into four "classes" corresponding to the number of "like" (blue) edge relations. This results in two balanced groups---3b and 1b---containing 3 and 1 "like" relationship respectively and two unbalanced groups (0b and 2b in Figure \ref{fig:triads}). In Figure \ref{fig:triads}, we provide examples of 4 individual triads sampled from each grouping. We note that we allow position to vary across triad images, as it should be irrelevant to this task.

\subsection{Component Graph Generation}

\begin{figure}[!h]
    \centering
    \includegraphics[scale=0.3]{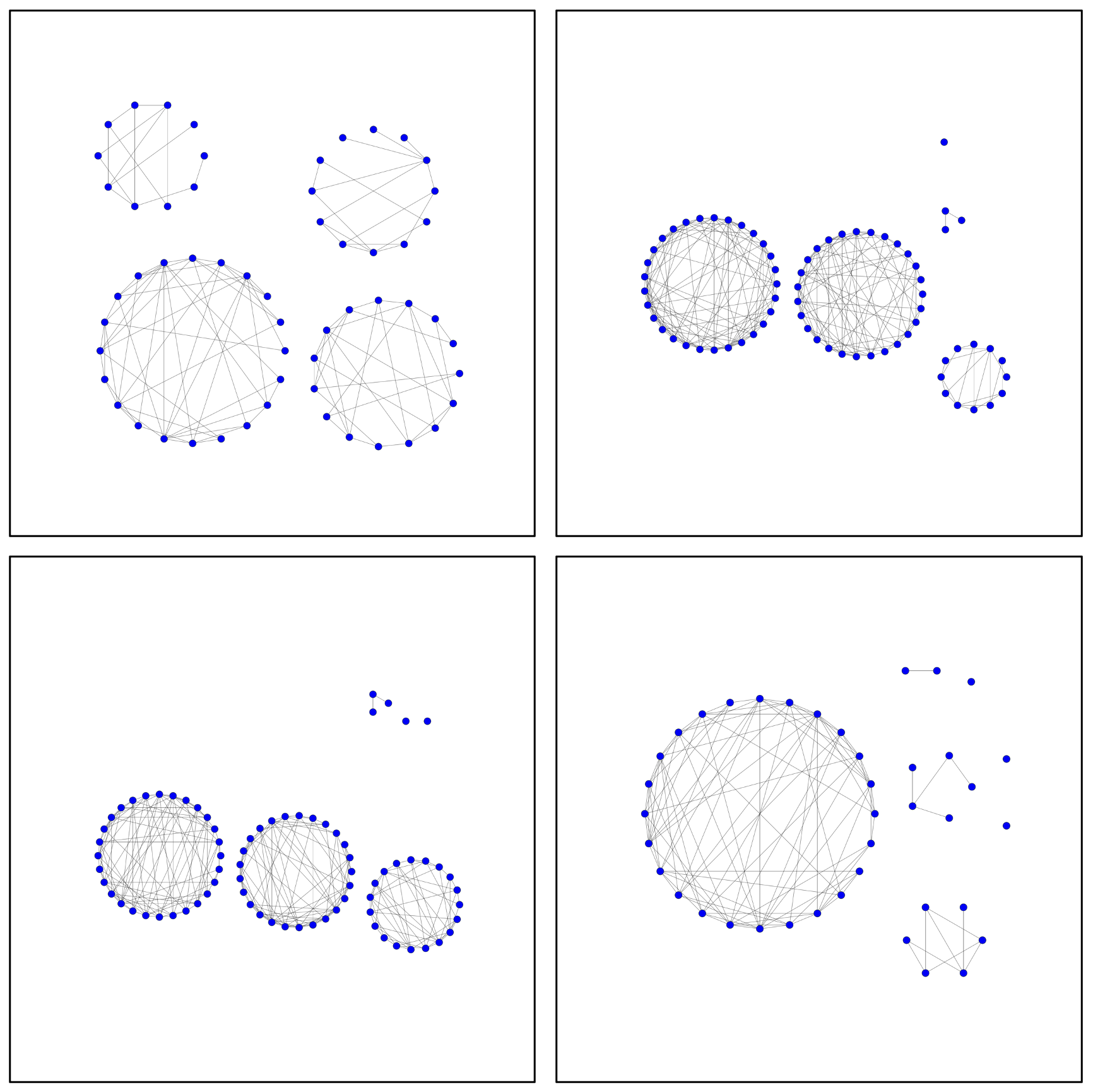}
    \caption{Components Example Graphs. Read from left to right and top to bottom, these graphs contain 4, 5, 6, and 7 components respectively. The graphs contain 0, 1, 2, and 3 isolates.}
    \label{fig:components}
\end{figure}

For each component graph, we independently generate 4 Erdos-Renyi graphs, with $p=0.3$ and where the number of nodes for each of the 4 graphs are randomly drawn integers between 0 and 30 inclusive. We then take the disjoint union between these four graphs, record the number of components and isolates in each graph, and visualize the result. Optimizing for human readability, we elected to visualize these graphs using the netgraph library, as it is optimized to support visual layouts containing multiple components \cite{Brodersen2023}. We chose a small-world layout for each component again for human readability. Component counts across all graphs ranged from 2 to 11 and isolate counts ranged from 0 to 6. We provide four example component graphs in Figure \ref{fig:components}. 

\section{Results}

\subsection{Maximum Degree Task Results}

We evaluate GPT-4 and LLaVa at both 1) identifying the maximum node degree in a graph and 2) identifying node IDs that have the maximal degree. Despite being able to define degree centrality, LLaVa performed extremely poorly on both tasks. On the "formal" prompt, where we instruct LLaVa to identify nodes the maximum degree and identify all nodes that have the maximal degree, LLaVa failed to correctly identify any maximum degrees. On the human prompts, it exhibited bizarre behavior; it frequently made up its own graphs in its responses and then (often incorrectly) answered the question based on the graph it generated. In both scenarios, its predictions often deviated substantially from ground truth, even with easy examples, as can be seen in the relatively high MSE scores in Table \ref{tbl:centrality}. In the top row of graphs in Figure \ref{fig:degree}, which very clearly has a maximum degree of one shared by two nodes, LLaVa's 4 runs identified maximum degrees of 3, 4, 10, and 11 respectively.

GPT-4 performed better than LLaVa on all metrics, but still struggled despite the simplicity of the task. A human-centric prompt with Letter IDs performed the best, yielded the best accuracy in identifying maximum degree, but did not have the best Mean Squared Error (MSE) nor the best Mean Jaccard Similarity. The human-centric prompt with letter IDs was the only run where GPT-4 correctly predicted the maximum degree of the graph in the top left corner of Figure \ref{fig:degree}. In all other runs, GPT-4 incorrectly identified the maximum degree of the top row graphs as 2.
\begin{table}[!h]
\centering
\begin{tabular}{ccc|cc|c|}
\cline{4-6}
                                             &                                              &             & \multicolumn{2}{c|}{Max Degree}     & Degree IDs \\ \cline{4-6} 
                                             &                                              &             & \multicolumn{1}{c|}{\textbf{Accuracy}} & \textbf{MSE} & \textbf{Mean Jaccard Similarity}    \\ \hline
\multicolumn{1}{|c|}{\multirow{4}{*}{gpt-4}}  & \multicolumn{1}{c|}{\multirow{2}{*}{Formal Prompt}} & Numeric IDs & \multicolumn{1}{c|}{0.45}     & \textbf{17 } & 0.491      \\ \cline{3-6} 
\multicolumn{1}{|c|}{}                       & \multicolumn{1}{c|}{}                        & Letter IDs  & \multicolumn{1}{c|}{0.35}     & 24  & 0.53       \\ \cline{2-6} 
\multicolumn{1}{|c|}{}                       & \multicolumn{1}{c|}{\multirow{2}{*}{Human Prompt}}  & Numeric IDs & \multicolumn{1}{c|}{0.4}      & 23  & \textbf{0.54}       \\ \cline{3-6} 
\multicolumn{1}{|c|}{}                       & \multicolumn{1}{c|}{}                        & Letter IDs  & \multicolumn{1}{c|}{\textbf{0.533}}    & 25  & 0.533      \\ \hline
\multicolumn{1}{|c|}{\multirow{4}{*}{LLaVa}} & \multicolumn{1}{c|}{\multirow{2}{*}{Formal Prompt}} & Numeric IDs & \multicolumn{1}{c|}{0}        & 693 & 0.075      \\ \cline{3-6} 
\multicolumn{1}{|c|}{}                       & \multicolumn{1}{c|}{}                        & Letter IDs  & \multicolumn{1}{c|}{0}        & 328 & 0.195      \\ \cline{2-6} 
\multicolumn{1}{|c|}{}                       & \multicolumn{1}{c|}{\multirow{2}{*}{Human Prompt}}  & Numeric IDs & \multicolumn{1}{c|}{0.15}     & 760 & 0.065      \\ \cline{3-6} 
\multicolumn{1}{|c|}{}                       & \multicolumn{1}{c|}{}                        & Letter IDs  & \multicolumn{1}{c|}{0.1}      & 572 & 0.054      \\ \hline
\end{tabular}
\label{tbl:centrality}
\caption{Results of GPT-4 and LLava on Identifying the maximum degree (Accuracy, MSE), and at identifying the set of nodes IDs that have the maximum degree in the graph (mean Jaccard similarity over all graphs).}
\end{table}

\subsection{Structural Balance Task Results}

For this balanced binary classification task, we categorize triads into four conceptual types, based on the number of ($+$) edges present in the triad, which we visualize in Figure \ref{fig:triads} and report accuracies for each subgroup in Table \ref{tbl:triads}. The highest overall accuracy attained was 0.51, on par with random guessing. However, performance was not identical on every subgroup.

GPT-4 generally performed well on the cases with 3 ($+$) edges (b1 in Figure \ref{fig:triads}) and 0 ($+$) edges (b0), achieving accuracies greater than or equal to 0.7 for balanced and unbalanced triads in these categories. However, it did surprisingly poorly across cases with 1 or 2 ($+$) edges. Balanced triads with 1 edge (b1) with and without a definition in the prompt received accuracies of 0.2 and 0.5, respectively. Unbalanced triads with 2 ($+$) edges (2b) received accuracies of 0.467 and 0.367 respectively. We note that GPT-4's reasoning was often inconsistent or faulty, even when a clear definition of structural balance was provided in the prompt. For example, on one of the (b0) categories that it incorrectly classified as balanced, GPT-4 returned the justification: ``This triad is balanced as all three edges depict "dislike" relationships (even number of "dislike", odd number of "like")''.

When provided with a clear definition of what constitutes structural balance, LLaVa predicted that every triad was unbalanced. Without a definition, LLaVa still largely overwhelmingly predicting that triads were unbalanced. In the cases where LLaVa predicted that triads were balanced, it generally offered a perplexing or inaccurate reasoning. For example, in several cases LLaVa stated that triads were balanced because "The blue and red lines are of equal length, indicating a balance between the two relationships". One of its most frequent (faulty) justifications for classifying a triad as unbalanced, accurately or inaccurately, was some variation of "the blue and red lines are not parallel, indicating an imbalance in the relationships."

\begin{table}[!h]
\centering
\begin{tabular}{cc|cccccc|c|}
\cline{3-9}
                                             &                                 & \multicolumn{7}{c|}{Accuracy}     \\ \cline{3-9} 
                                             &                                 & \multicolumn{1}{c|}{\textbf{3b}}  & \multicolumn{1}{c|}{\textbf{1b}}    & \multicolumn{1}{c|}{\textbf{b0}}  & \multicolumn{1}{c|}{\textbf{2b}}    & \multicolumn{1}{c|}{\textbf{balanced}} & \textbf{unbalanced} & \textbf{overall} \\ \hline
\multicolumn{1}{|c|}{\multirow{2}{*}{gpt-4}}  & No Definitions & \multicolumn{1}{c|}{0.70} & \multicolumn{1}{c|}{\textbf{0.50}}   & \multicolumn{1}{c|}{0.80} & \multicolumn{1}{c|}{0.37} & \multicolumn{1}{c|}{\textbf{0.55}}       & 0.46       & \textbf{0.51}  \\ \cline{2-9} 
\multicolumn{1}{|c|}{}                       & Definitions                     & \multicolumn{1}{c|}{\textbf{1}}   & \multicolumn{1}{c|}{0.20}   & \multicolumn{1}{c|}{0.90} & \multicolumn{1}{c|}{0.47} & \multicolumn{1}{c|}{0.40}        & 0.58       & 0.49  \\ \hline
\multicolumn{1}{|c|}{\multirow{2}{*}{LLaVa}} & No Definitions & \multicolumn{1}{c|}{0.10} & \multicolumn{1}{c|}{0.23} & \multicolumn{1}{c|}{0.80} & \multicolumn{1}{c|}{0.73} & \multicolumn{1}{c|}{0.20}        & 0.75       & 0.48   \\ \cline{2-9} 
\multicolumn{1}{|c|}{}                       & Definitions                         & \multicolumn{1}{c|}{0}   & \multicolumn{1}{c|}{0}     & \multicolumn{1}{c|}{\textbf{1}}   & \multicolumn{1}{c|}{\textbf{1}}     & \multicolumn{1}{c|}{0}          & \textbf{1}          & 0.50     \\ \hline
\end{tabular}
\label{tbl:triads}
\caption{Triadic Balance Results: b denotes the number of 'like' (blue) edges. 3b and 0b report accuracy only on triads containing 3 and 0 ($+$) relations respectively. 1b and 2b contain accuracy metrics on the triads containing 1 and 2 ($+$) relations respectively. The balanced column contains accuracy calculated on all balanced triads (3b and 1b), unbalanced is calculated on all unbalanced triads (b0 and 2b) and overall is accuracy calculated over all triads.}
\end{table}

\subsection{Component Task Results}

The top performing model across all evaluation metrics for both the component and isolate counting tasks was GPT-4 with no definitions included in the prompt. However, we note that the inclusion of definitions resulted in a substantial improvement of LLaVa's MSE, which decreased from over 20,000 to below 11. gpt-4 performed relatively well on the isolate counting task, and provided the correct number of isolates 67 of 100 times. We provide results in Table \ref{tbl:components}.

\begin{table}[]
\centering
\begin{tabular}{cc|ccc|ccc|}
\cline{3-8}
                                             &                & \multicolumn{3}{c|}{Components}                                                         & \multicolumn{3}{c|}{Isolates}                                                           \\ \cline{3-8} 
                                             &                & \multicolumn{1}{c|}{Accuracy}      & \multicolumn{1}{c|}{MAE}           & MSE           & \multicolumn{1}{c|}{Accuracy}      & \multicolumn{1}{c|}{MAE}           & MSE           \\ \hline
\multicolumn{1}{|c|}{\multirow{2}{*}{gpt-4}}  & No Definitions & \multicolumn{1}{c|}{\textbf{0.39}} & \multicolumn{1}{c|}{\textbf{1.25}} & \textbf{3.41} & \multicolumn{1}{c|}{\textbf{0.67}} & \multicolumn{1}{c|}{\textbf{0.51}} & \textbf{0.99} \\ \cline{2-8} 
\multicolumn{1}{|c|}{}                       & Definitions    & \multicolumn{1}{c|}{0.37}          & \multicolumn{1}{c|}{1.45}          & 4.97          & \multicolumn{1}{c|}{0.64}          & \multicolumn{1}{c|}{0.58}          & 1.3           \\ \hline
\multicolumn{1}{|c|}{\multirow{2}{*}{LLaVa}} & No Definitions & \multicolumn{1}{c|}{0.03}          & \multicolumn{1}{c|}{59.79}         & 23297.95      & \multicolumn{1}{c|}{0.08}          & \multicolumn{1}{c|}{32.43}         & 20652.51      \\ \cline{2-8} 
\multicolumn{1}{|c|}{}                       & Definitions    & \multicolumn{1}{c|}{0.05}          & \multicolumn{1}{c|}{2.81}          & 10.57         & \multicolumn{1}{c|}{0.17}          & \multicolumn{1}{c|}{1.27}          & 2.23          \\ \hline
\end{tabular}
\label{tbl:components}
\caption{Component Results. We report Accuracy, MAE, and MSE for each model and prompt condition for both the Component counting and Isolate counting tasks.}
\end{table}

\section{Discussion and Limitations}

Given GPT-4's strong performance on professional exams like the LSAT and MCAT, it is, at least from a human perspective, surprising that it would struggle with something as simple as counting specific elements of graphs. This phenomenon likely relates, at least in part, to how LVMs process images as patches \cite{xu2023zero}. Nonetheless, more research is needed to understand why LVMs struggle on tasks this simple, and future research could explore the performance of LVMs fine-tuned on graph-related tasks.

A limitation of this study is that there are a large number of ways to visualize any given graph, and there are myriad parameters that are chosen which may impact LVM evaluation. It's possible that something as trivial as changing node color could impact LVM evaluation. While the tasks we designed are not representative of all possible graphs, they do evaluate mastery and understanding of the underlying network science concepts. The ways that graph visualization parameter selection and prompt engineering impact LVM performance on VNA tasks are clear and important avenues for future research.

\section{Conclusion}

We propose the task of zero-shot Visual Network Analysis to evaluate the performance of LVMs on graph analytics tasks. We create a benchmark that includes 5 tasks related to 3 core network science concepts: maximum degree, structural balance, and identifying components. We find that across all tasks, LLMs struggled to identify and count the appropriate element of graphs---an essential skill in analyzing network data. We publicly release all generated data and ground-truth labels.

\begin{credits}
\subsubsection{\ackname} 

The research for this paper was supported in part by the ARMY Scalable Technologies for Social Cybersecurity and the Office of Naval Research, MURI: Persuasion, Identity, \& Morality in Social-Cyber Environments under grants W911NF20D0002 and N000142112749. It was also supported by the center for Informed Democracy and Social-Cybersecurity  (IDeaS) and the center for Computational Analysis of Social and Organizational Systems (CASOS) at Carnegie Mellon University. The views and conclusions  are those of the authors and should not be interpreted as representing the official  policies, either expressed or implied, of the ARMY, the ONR, or the US Government.

\subsubsection{\discintname}
The authors have no competing interests to declare that are relevant to the content of this article.
\end{credits}
%
%
%
\bibliographystyle{splncs04}
\bibliography{llm_vna}

\begin{thebibliography}{10}
\providecommand{\url}[1]{\texttt{#1}}
\providecommand{\urlprefix}{URL }
\providecommand{\doi}[1]{https://doi.org/#1}

\bibitem{achiam2023gpt}
Achiam, J., Adler, S., Agarwal, S., Ahmad, L., Akkaya, I., Aleman, F.L., Almeida, D., Altenschmidt, J., Altman, S., Anadkat, S., et~al.: Gpt-4 technical report. arXiv preprint arXiv:2303.08774  (2023)

\bibitem{Brodersen2023}
Brodersen, P.J.N.: Netgraph: Publication-quality network visualisations in python. Journal of Open Source Software  \textbf{8}(87), ~5372 (2023). \doi{10.21105/joss.05372}, \url{https://doi.org/10.21105/joss.05372}

\bibitem{cartwright1956structural}
Cartwright, D., Harary, F.: Structural balance: a generalization of heider's theory. Psychological review  \textbf{63}(5), ~277 (1956)

\bibitem{chen2024exploring}
Chen, Z., Mao, H., Li, H., Jin, W., Wen, H., Wei, X., Wang, S., Yin, D., Fan, W., Liu, H., et~al.: Exploring the potential of large language models (llms) in learning on graphs. ACM SIGKDD Explorations Newsletter  \textbf{25}(2),  42--61 (2024)

\bibitem{fatemi2023talk}
Fatemi, B., Halcrow, J., Perozzi, B.: Talk like a graph: Encoding graphs for large language models. arXiv preprint arXiv:2310.04560  (2023)

\bibitem{goh2021multimodal}
Goh, G., †, N.C., †, C.V., Carter, S., Petrov, M., Schubert, L., Radford, A., Olah, C.: Multimodal neurons in artificial neural networks. Distill  (2021). \doi{10.23915/distill.00030}, https://distill.pub/2021/multimodal-neurons

\bibitem{harary1953notion}
Harary, F.: On the notion of balance of a signed graph. Michigan Mathematical Journal  \textbf{2}(2),  143--146 (1953)

\bibitem{heider1946attitudes}
Heider, F.: Attitudes and cognitive organization. The Journal of psychology  \textbf{21}(1),  107--112 (1946)

\bibitem{islam2023pushing}
Islam, A., Biswas, M.R., Zaghouani, W., Belhaouari, S.B., Shah, Z.: Pushing boundaries: Exploring zero shot object classification with large multimodal models. In: 2023 Tenth International Conference on Social Networks Analysis, Management and Security (SNAMS). pp.~1--5. IEEE (2023)

\bibitem{li2024visiongraph}
Li, Y., Hu, B., Shi, H., Wang, W., Wang, L., Zhang, M.: Visiongraph: Leveraging large multimodal models for graph theory problems in visual context. arXiv preprint arXiv:2405.04950  (2024)

\bibitem{liu2024llavanext}
Liu, H., Li, C., Li, Y., Li, B., Zhang, Y., Shen, S., Lee, Y.J.: Llava-next: Improved reasoning, ocr, and world knowledge (January 2024), \url{https://llava-vl.github.io/blog/2024-01-30-llava-next/}

\bibitem{wasserman1994social}
Wasserman, S., Faust, K.: Social network analysis: Methods and applications  (1994)

\bibitem{wei2024rendering}
Wei, Y., Fu, S., Jiang, W., Kwok, J.T., Zhang, Y.: Rendering graphs for graph reasoning in multimodal large language models. arXiv preprint arXiv:2402.02130  (2024)

\bibitem{xu2023zero}
Xu, J., Le, H., Samaras, D.: Zero-shot object counting with language-vision models. arXiv preprint arXiv:2309.13097  (2023)

\bibitem{zeng2024large}
Zeng, J., Huang, R., Malik, W., Yin, L., Babic, B., Shacham, D., Yan, X., Yang, J., He, Q.: Large language models for social networks: Applications, challenges, and solutions. arXiv preprint arXiv:2401.02575  (2024)

\bibitem{zhang2024good}
Zhang, C., Wang, S.: Good at captioning, bad at counting: Benchmarking gpt-4v on earth observation data. arXiv preprint arXiv:2401.17600  (2024)

\bibitem{zhang2024vision}
Zhang, J., Huang, J., Jin, S., Lu, S.: Vision-language models for vision tasks: A survey. IEEE Transactions on Pattern Analysis and Machine Intelligence  (2024)

\end{thebibliography}

\end{document}